\documentclass[9pt, conference]{IEEEtran}
\IEEEoverridecommandlockouts
\usepackage[utf8]{inputenc}
\usepackage{cite}
\usepackage{amsmath,amssymb,amsfonts}
\usepackage{algorithmic}
\usepackage{graphicx}
\usepackage{textcomp}
\usepackage{xcolor}
\usepackage{makecell}
\usepackage{xspace}%
\usepackage{multirow}
\usepackage{booktabs}
\usepackage{stfloats}
\usepackage{url}
\usepackage{float}

\newcommand\modelname[0]{\emph{SDXL}\xspace}

\newcommand\sdshort[0]{\emph{SD}\xspace}

\usepackage{array}
\def\BibTeX{{\rm B\kern-.05em{\sc i\kern-.025em b}\kern-.08em
    T\kern-.1667em\lower.7ex\hbox{E}\kern-.125emX}}
\begin{document}

\title{Enhancing Image Generation Fidelity via Progressive Prompts\\
}

\author{
\IEEEauthorblockN{Zhen Xiong$^{1}$, Yuqi Li$^{1}$\thanks{Zhen Xiong \&Yuqi Li are interns}, Chuanguang Yang$^{1}$, Tiao Tan$^{2}$, Zhihong Zhu$^{3}$, Siyuan Li$^{4}$, Yue Ma$^{5}$\dag\thanks{\IEEEauthorrefmark{2}Corresponding author, Email: mayuefighting@gmail.com }}
\IEEEauthorblockA{
\textit{$^1$Institute of Computing Technology, Chinese Academy of Sciences, China}\\
\textit{$^2$Tsinghua University, China}\\
\textit{$^3$Peking university, China}\\
\textit{$^4$EaseUS, China}\\
\textit{$^5$The Hong Kong University of Science and Technology, HK}\\
}
}

\maketitle

\begin{abstract}
Diffusion transformer (DiT) architecture catches much attention in image generation, which achieves better fidelity, performance, and diversity.
However, most existing DiT-based image generation methods are global-aware synthesis and regional prompt control is less explored. In this paper, we propose a coarse-to-fine generation pipeline for regional prompt-following generation. 
Specifically, we first leverage the powerful large language model (LLM) to generate the high-level description of image (such as content, topic, and objects) and low-level description of image (such as details and style). Then we explore the influence of cross-attention layers in different depths. We discover that deeper layers always responsible for the high-level content control, while the shallow layers handles low-level content control.
The various prompts are injected into the proposed regional cross-attention control in order for course-to-fine generation. Using the proposed pipeline, we improve the controllability of DiT-based image generation. Extensive quantitative and qualitative results demonstrate that our pipeline enables to improve the generated performance. Our codes are available at \url{https://github.com/ZhenXiong-dl/ICASSP2025-RCAC}.

\end{abstract}


\section{Introduction}
Recent development of diffusion models~\cite{ho2020denoising, podell2023sdxl,rombach2022high, clip, yang2024clip, feng2024relational} has improved the performance of text-to-image generation, such as Stable Diffusion XL~\cite{podell2023sdxl}, DALL-E 3~\cite{betker2023improving}, and Imagen~\cite{saharia2022photorealistic}. Due to the ability to scale up,  researchers have begun to explore how to use the diffusion transformer (DiT) as the backbone, which is much more faster, impressive, and realistic. Several works leverage the DiT to push the image generation to a new peak, such as hunyuan-dit~\cite{li2024hunyuan}, pixart~\cite{chen2023pixart}, and so on. Even though
their remarkable ability to synthesize realistic images
content with text prompts, DiT-based T2I generation~\cite{li2024hunyuan, wang2024gra, chen2023pixart, glide, transformer, clipscore} struggles to generate the detailed image using complex prompt guidance, which describes the style, texture, and color. 

Some works~\cite{zhang2023adding, li2023gligen, bar2023multidiffusion, wang2024taming, zhu2024multibooth, dino,yang2022cross, li2024sglp, li2024comae, yang2024eva, li2025fedkd} address this challenge by additional condition guidance, including canny, box, and layout. Equipping the prompt-aware attention guidance, they enable to improve compositional T2I generation. For instance, GLIGEN leverages the proposed self-attention strategy to incorporate spatial domain, while freezing the original weight to maintain the powerful generation ability.  GORS uses the highly image-prompt aligned generated dataset to fine-tune pre-trained text-to-image model and apply text-image alignment reward to balance loss. Additionally, imageReward sets up a general-purpose reward model to improve the ability of prompt alignment. However, these approaches are all based on the UNet architecture and few works explore the complex prompt following in DiT-based image generation.

\begin{figure}[t]  
	\centering
	\includegraphics[width=0.85\linewidth, height=0.85\linewidth]{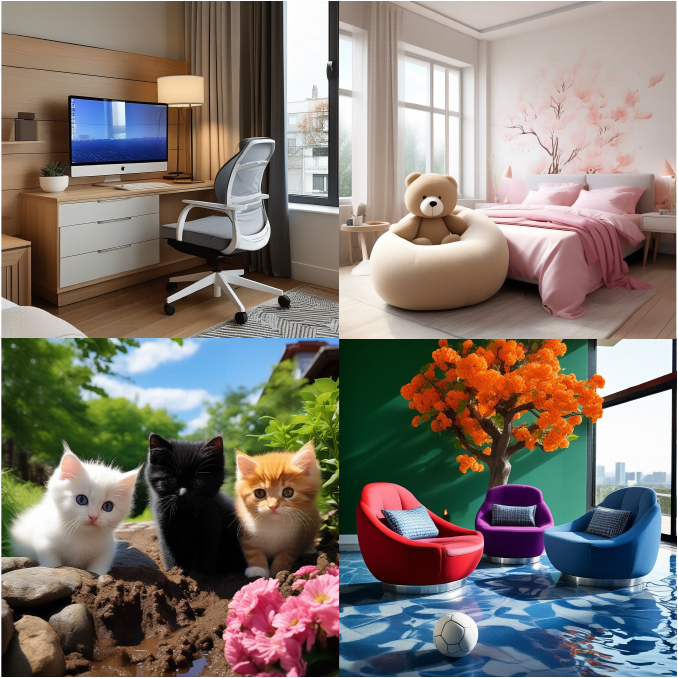}
	\caption{Visual results of the proposed approach. We enable to generate the image with more details, such as lighting, and objects.}
	\label{fig1}
	\vspace{-0.5cm}
\end{figure}

In order to improve the fidelity of generated results, we propose DiTPipe, a 
coarse-to-fine generation pipeline for regional prompt-following generation. In particular, we first produce high-level prompts and low-level prompts using the powerful large language model (LLM). High-level prompt always includes the description about layout and style, while low-level prompts focus on the details of color, texture, and object. Then, we study the influence of cross-attention layers in different depths of diffusion transformer and propose a regional cross-attention control strategy to enhance the generated details in different regions. We perform extensive quantitative and qualitative results to proof the superiority of proposed approaches. Compared with previous work~\cite{podell2023sdxl, ma2024followyourpose, ma2024followyourclick, ma2022visual, ma2023magicstick, ma2024followyouremoji, chen2024follow, wang2024cove, zhu2024instantswap, feng2024dit4edit, li2024hunyuan, t5, db, ipada, ma2024discrepancy, lu2024generic}, our approach achieves better performance.
To summarize, our main contributions are as follows:

\begin{itemize}

\item In order to improve the complex prompt-following ability of DiT~\cite{peebles2023scalable}, we propose DiTPipe, which is a coarse-to-fine generation pipeline for the regional prompt-following generation.

\item We propose progressive-prompt image generation, which includes high-level prompts and low-level prompts. We also design a regional cross-attention control strategy to enhance the fidelity of generated images.

\item Extensive quantitative and qualitative experiments demonstrate that our proposed pipeline achieves better performance 

\end{itemize}


\section{Methodology}
\begin{figure}[t]  
	\centering
	\includegraphics[width=1.0\linewidth]{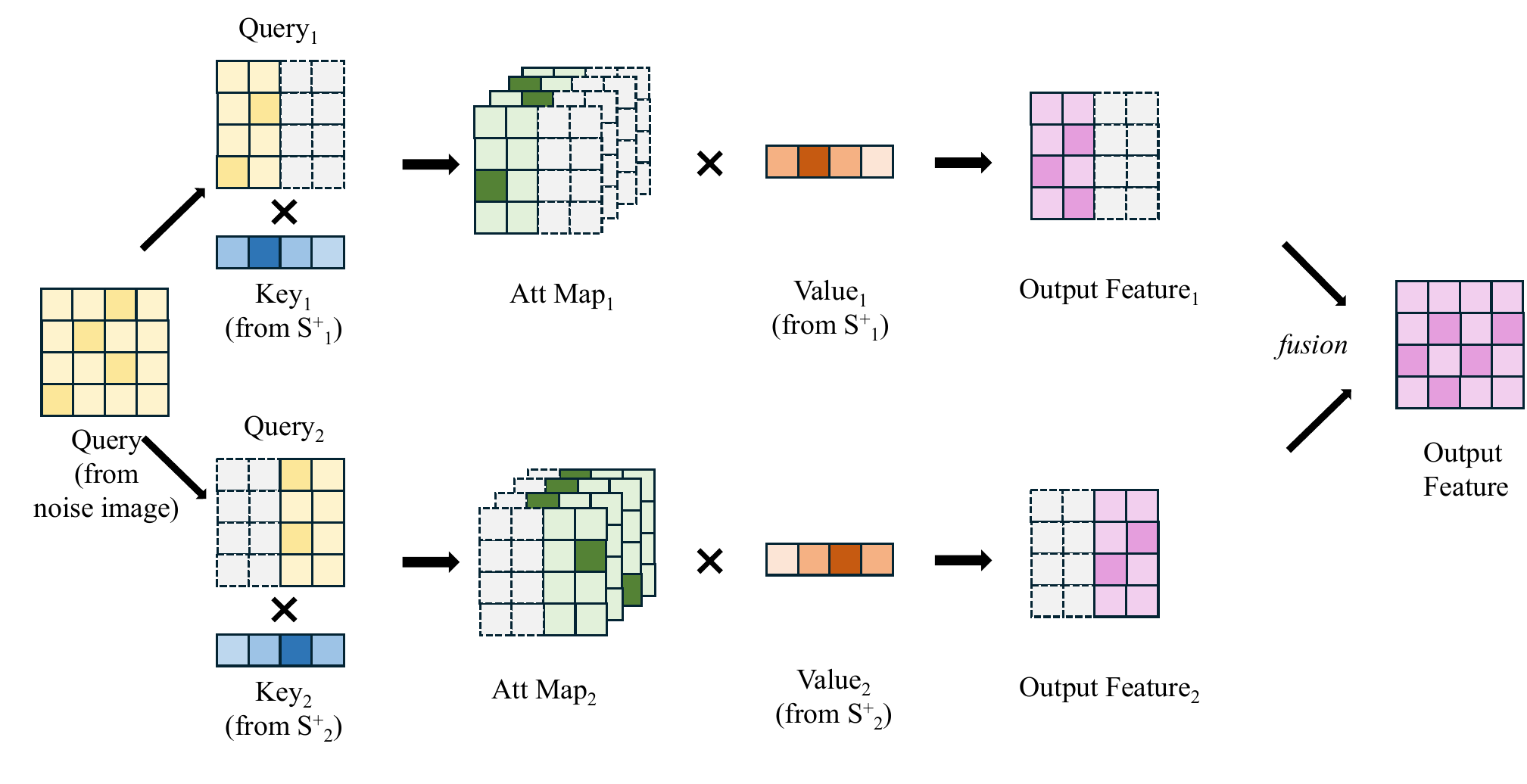}
	\caption{The details of proposed region-attention. We present the control pipeline of two regions. Different prompts are applied for specific area guidance. Then, we fuse them to final representation.}
	\label{fig2}
	\vspace{-0.5cm}
\end{figure}


\begin{figure}[t]  
	\centering
	\includegraphics[width=1.0\linewidth]{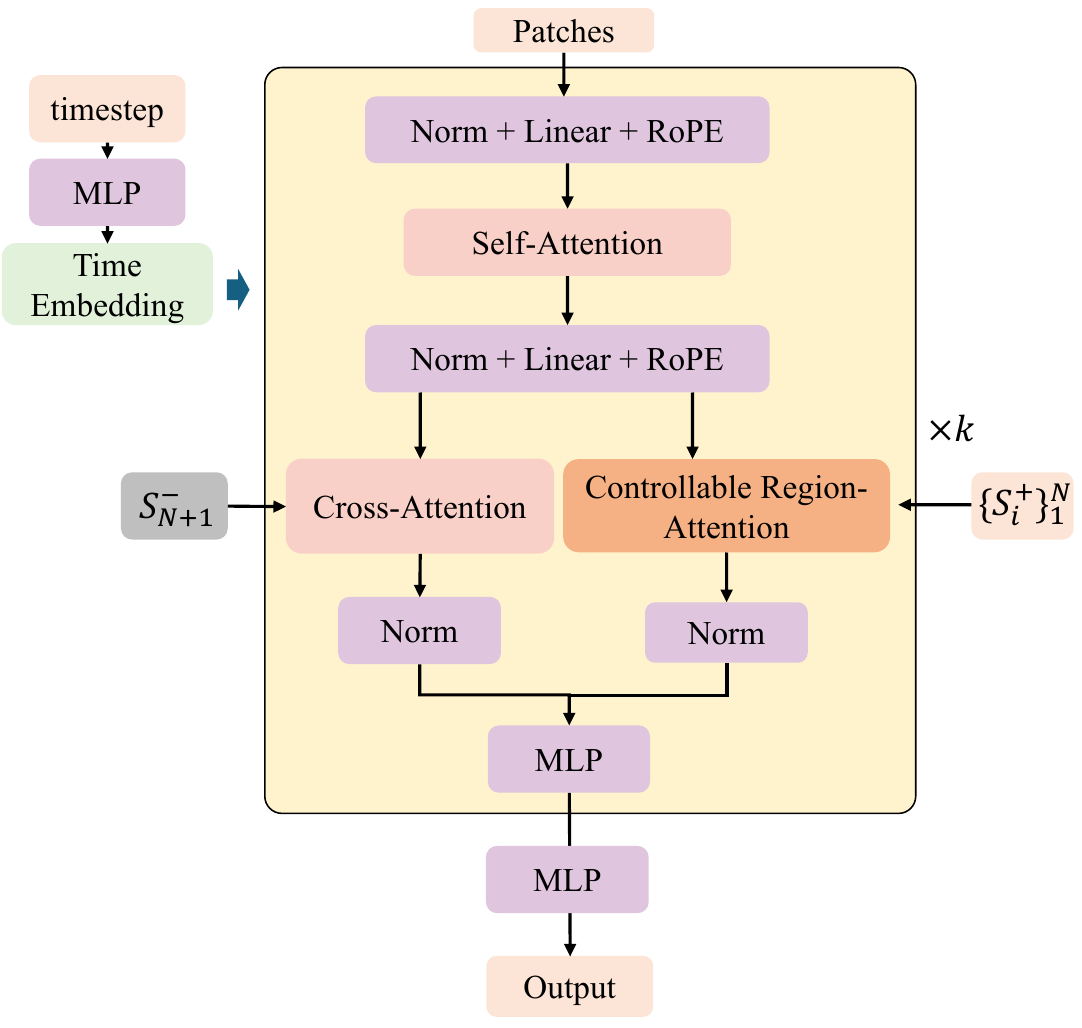}
	\caption{The details of our proposed block. In the figure, we present the modified block of DiT. In order to improve the fidelity of results, we design the Controllable Region-Attention to improve to achieve more accurate control.}
	\label{fig3}
	\vspace{-0.5cm}
\end{figure}

\begin{figure}[t]  
	\centering
	\includegraphics[width=0.7\linewidth]{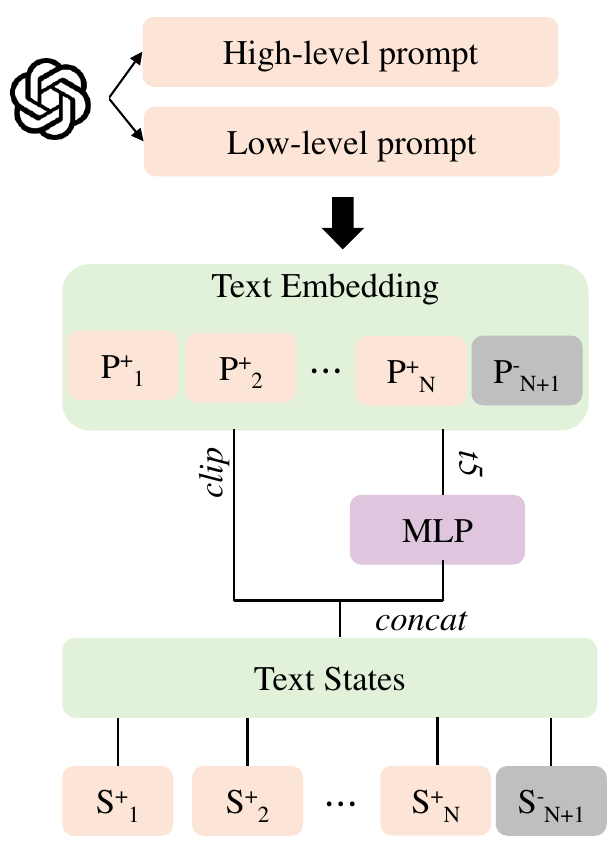}
	\caption{The process of prompt generation, we show the process of progressive prompts, including high-level prompts and low-level prompts.}
	\label{fig4}
	\vspace{-0.5cm}
\end{figure}

The Diffusion Transformers (DiT) architecture introduces the powerful modeling capabilities of transformers into the visual domain by integrating them with the diffusion framework. Specifically, DiT replaces the traditional U-Net network with DiT blocks, enabling higher-quality image generation. In particular, the cross-attention within the DiT block introduces the text conditions and image latent to participate in noise. Modality fusion establishes a foundation for the application of distinct prompts to designated locations within the image latent space, enabling precise region-specific control. Building on these observations, we propose a method called Controllable Region-Attention. Specifically, this method injects different textual (prompt) features into multiple local regions of the image representation within the cross-attention layers of the DiT architecture. Leveraging the strong text comprehension capabilities of the T5 encoder and the high degree of image-text coupling enabled by the substantial number of Dit Blocks through cross-attention mechanisms, this approach enables more precise control over the representation of image features in different regions.

Our Controllable Region-Attention method focuses solely on modifying the attention mechanism without depending on other specific components of the model, making it a fast and universally adaptable plug-and-play solution for diffusion models based on the Transformer architecture, such as those utilizing the DiT structure. Since our approach is built upon the Hunyuan-DiT Block, the T5-encoded text features are seamlessly segmented and injected into the local image features, ensuring better alignment between the local image regions and the corresponding textual control instructions.

\subsection{Region Mask Division and Prompts Setting}
First, given an image $I$, we divide it along either the height or width dimension into $N$ adjacent controllable regions, $R_1, R_2, \dots, R_N$. This division allows us to localize specific areas within the image that can be independently controlled during the generation process. The choice of the dimension (height or width) for division is dependent on the desired granularity and the nature of the image content, which can be tailored based on the specific application or task at hand. 

Next, we configure $N+1$ prompts ($P_1^+, P_2^+, \dots, P_N^+, P_{N+1}^-$), where the first $N$ prompts, $P_{i}^+$, are applied as positive conditions to region $R_{i}$. This approach ensures that each region of the image can be guided by a distinct semantic prompt, thereby enabling fine-grained control over the generated content. The prompts can be designed to reflect specific attributes or concepts that the user wishes to emphasize within each region. For example, in a landscape image, one region might be controlled to generate a sky with specific weather conditions, while another region might be controlled to depict a particular type of terrain.

The final prompt, $P_{N+1}^-$, is applied as a global negative condition to the whole image $I$. The inclusion of a negative prompt is crucial as it acts as a constraint, helping to suppress undesired features or attributes that may otherwise emerge in the generated image. This negative prompt also promotes a more balanced and unified generation style across different controllable regions.
\begin{figure*}[!b]  
	\centering
	\includegraphics[width=\linewidth]{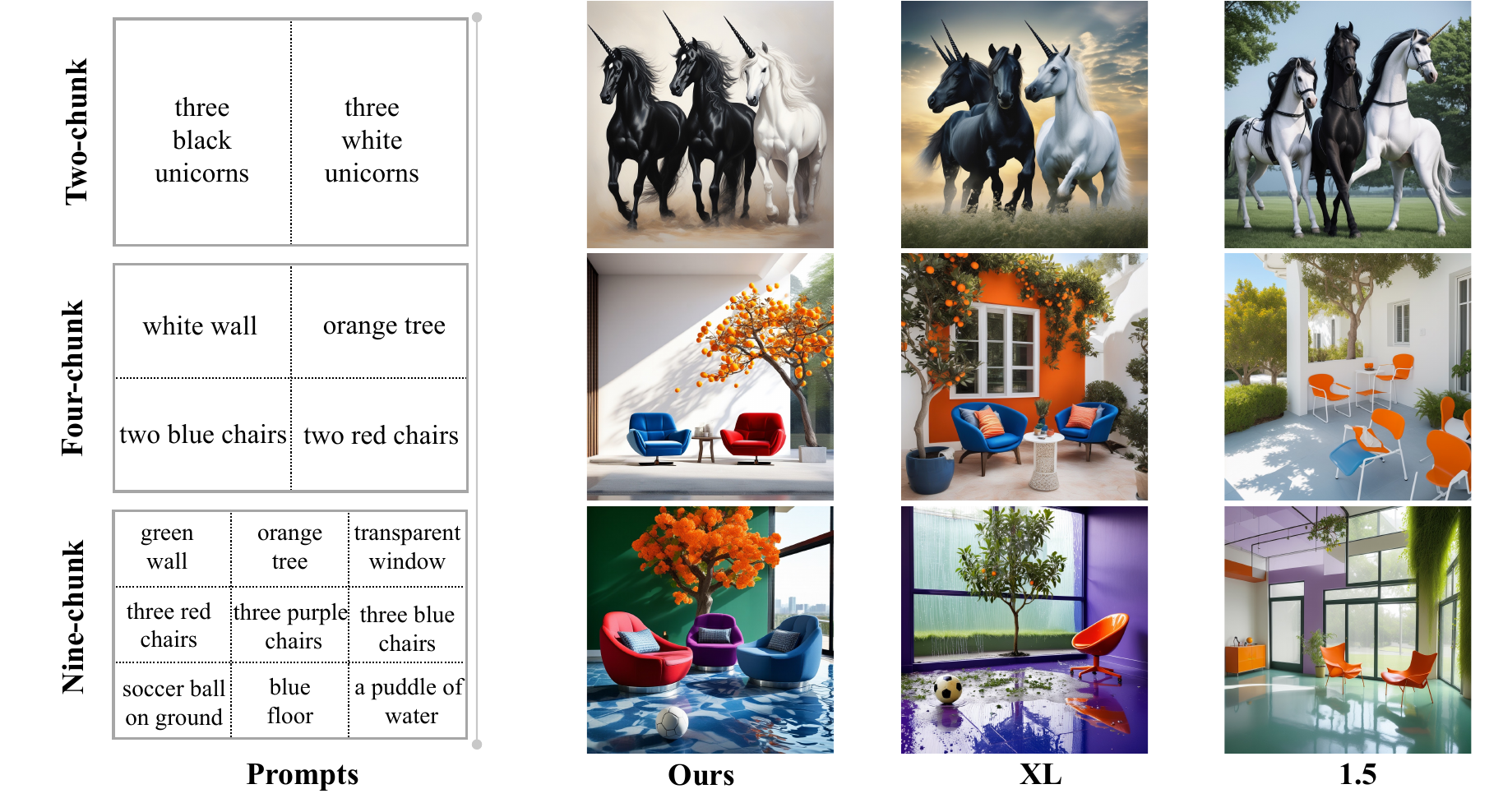}
	\caption{Comparison results. We show the comparison with SDXL and SD-1.5. The left of figure is the low-level prompts. We perform the experiment on three setting, including two, four, and nine chunks. Our pipeline obtain the better performance.}
	\label{fig5}
	\vspace{-0.5cm}
\end{figure*}

\subsection{Text Embedding Preprocessing}
In our implementation, given that we utilize Hunyuan’s cross-attention DiT architecture, it is necessary to first pass the T5 embeddings for each prompt through an MLP layer for transformation. Subsequently, the T5 embeddings are concatenated with the CLIP embeddings along the sequence length dimension to produce the required 333-length text states for the Hunyuan-DiT block. In practice, we group all the positive prompts for the multiple regions and the global negative prompt into a single batch. This allows us to efficiently generate the text states required for the transformer block in one pass for all prompt embeddings, as illustrated in Figure 4.

\subsection{Attention Fusion}
Figure 2 illustrates the process by which multiple positive text states are propagated and fused within the Controllable Region Attention mechanism. First, for each controllable region $R_i$ (for simplicity, $i\leq 2$ in our illustration), a corresponding mask, denoted as $\text{Mask-Orig}_i$, is generated based on the respective region within the original image space. This mask is subsequently down-sampled to match the resolution of the latent image. Afterward, the mask is flattened along the height and width dimensions, resulting in $\text{Mask}_i$. For intuitive understanding, the Queries are visualized as 2D spatial representations in Figure 2, though they are flattened latent features in practice. 
\begin{figure*}[!t]  
	\centering
	\includegraphics[width=1.0\linewidth]{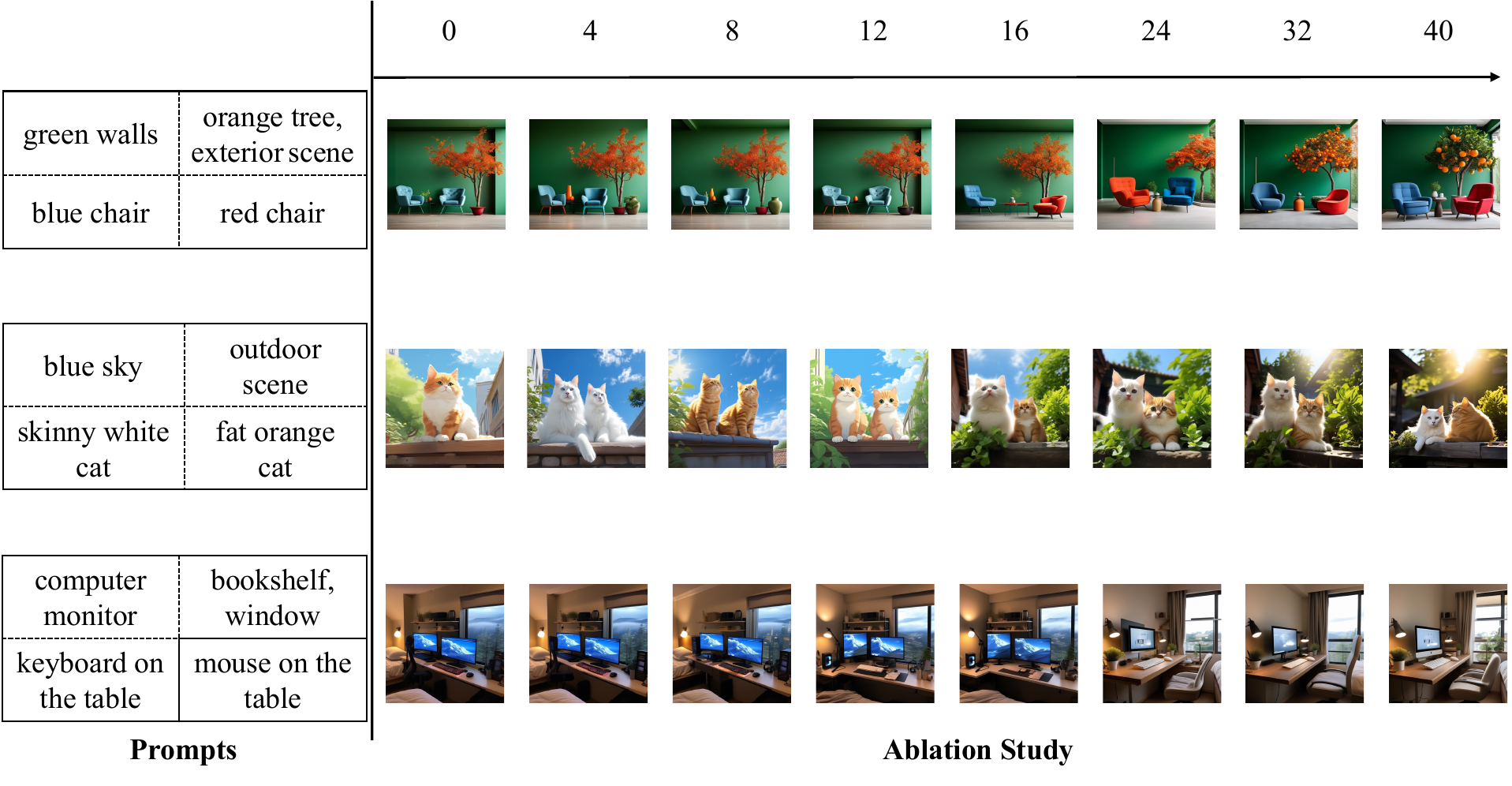}
	\caption{The ablation study of our method applied to different depth of the model while keeping the total number of layers \textbf{constant}. We inject the low-level prompts (\textit{Left}) into different depths and observe that the deeper cross-attention injection has better controllability (\textit{Right}).}
	\label{fig6}
	\vspace{-0.5cm}
\end{figure*}

Within the cross-attention mechanism, the Query is element-wise multiplied with the corresponding region mask $\text{Mask}_i$, effectively zeroing out the areas of the image outside of $R_i$. This operation ensures that only the features within the target region are retained for further processing. The masked Query, $\text{Query}_i$, is then matrix-multiplied with the $\text{Key}_i = \text{Linear}(S_i)$, where $S_i$ represents the text state associated with prompt $P_i^+$. This results in the computation of the attention map, denoted as $\text{Att Map}_i$. 

Subsequently, the attention map is used to compute the local output feature $f_i$ by multiplying it with the $\text{Value}_i = \text{Linear}(S_i)$. This produces the region-specific output feature that corresponds to the prompt applied to $R_i$. Finally, all local output features are aggregated across regions, resulting in the final output of the Controllable Region Attention module: $\text{Feature} = \sum_{i} f_i$.

For the negative prompt, we adopt a similar approach to the handling in SDXL. The text states corresponding to the negative prompt are processed independently in the separate cross-attention module and concatenated with the previous output generated by our Controllable Region Attention module along the batch dimension.

The above operations are applied at each DiT block throughout the network. In our architecture settings, this process is repeated 39 times, which ensures that the local prompts are accurately embedded into the corresponding areas of the latent feature space.

\section{experiment}
\subsection{Implementation Details}
To facilitate the injection of multiple controllable regional prompts, we employ the Hunyuan-DiT architecture as our base model and integrate the Controllable Region-Attention module. This module enables the alignment of image and text information across up to \( N \) positive prompts. In our experimental setup, we conducted extensive evaluations with \( N \in \{2, 4, 9\} \), covering a range of region-specific control scenarios. Meanwhile, the native cross-attention module remains responsible for integrating a global negative prompt, which serves to suppress the generation of undesired features or content in the image.

Our experiments were conducted on an NVIDIA RTX 4090 GPU. The resolution of images is configured as 1024\(\times\)1024 pixels. The model was trained using the SGM uniform scheduler. For sampling, we employed the Euler sampler and  SGM uniform scheduler. Additionally, the classifier-free guidance (CFG) scale was set to 6, and the initial denoise parameter was set to 1.

\begin{table}[!t]
\centering
\caption{\label{tab:aecomp} Comparison with previous work, our approach achieve better performance.}
\resizebox{.5\textwidth}{!}{%
\begin{tabular}{lcccc}
\toprule
model & PNSR $\uparrow$ & SSIM $\uparrow$ & LPIPS $\downarrow$ & rFID $\downarrow$ \\
\midrule
\modelname-VAE & 24.7 & 0.73 & 0.88 & 4.4 \\
\sdshort-VAE 1.x & 23.4 & 0.69 & 0.96 & 5.0 \\ 
\sdshort-VAE 2.x & 24.5 & 0.71 & 0.92 & 4.7 \\ 
\textbf{Ours} & \textbf{28.2} & \textbf{0.75} & \textbf{0.84} & \textbf{4.2}\\
\bottomrule
\end{tabular}}
\vspace{-1em}
\end{table}

\subsection{Baseline Comparison}
We compare the performance of our method against several baseline models, including the vanilla Hunyuan-DiT, SDXL (RealVisXL – v5), and SD-1.5 (majicMIX realistic – v7), for text-to-image generation tasks involving \( N \) controllable regions (Figure 5). Since these baseline models do not natively support region-specific prompt injection, we merge all prompts into a single global instruction and evaluate their performance on text-to-image generation. This baseline serves as a control to highlight the benefits of region-specific control enabled by our method.

We leverage the four matrixes: 1) PSNR: Measures the fidelity of generated images by comparing the signal-to-noise ratio, with higher values indicating better quality. 2) SSIM: Assesses structural similarity between images, focusing on perceptual aspects like luminance, contrast, and structure. 3) LPIPS: A perceptual metric using deep networks to evaluate visual similarity, where lower values indicate closer resemblance to reference images. 4) rFID: Measures the distribution similarity between generated and real images, with lower values indicating more realistic and diverse image generation.

Additionally, we compare our approach with the Couple method, which allows control over two distinct regions within an image. The Couple method can be viewed as a specific case of our pipeline when restricted to two regions. During the experiments, we evaluate both methods in terms of content control accuracy, spatial coherence, and semantic fidelity, as shown in Figure 5. These comparisons allow us to quantitatively assess the improvements brought by our method in scenarios requiring fine-grained regional control.

\subsection{Ablation Studies}
To further validate the contribution of the Controllable Region-Attention module, we conduct ablation studies to analyze its role in aligning specific image regions with their corresponding prompts. Firstly, we replace all of our Controllable Region-Attention modules with the standard cross-attention modules from the DiT architecture across the denoising pipeline. This enables us to isolate the impact of our proposed module on image generation. Then, we gradually increased the number of Controllable Region-Attention modules from none (0) to all (39) to present a gradual progression of the model's ability to adhere to regional prompts and control object placement. This incremental approach allows us to evaluate how the number of Controllable Region-Attention modules affects the model's performance in terms of regional specificity and image quality.

Through these processes, we generate a series of images that demonstrate two key findings: (1) our module significantly enhances the capability of the model regarding object placement and region-specific prompt alignment, and (2) it leads to more coherent and semantically consistent visual outputs, particularly in terms of multiple similar objects with different colors. These results are depicted in Figure 6.





\section{Conclusion}
In this paper, we propose a coarse-to-fine generation pipeline for regional prompt-following generation. Specifically, We first leverage the powerful large language model (LLM) to generate high-level image descriptions (content, topic, objects) and low-level details (style, color). Then we explore the influence of cross-attention layers at different depths. Extensive quantitative and qualitative results demonstrate the superiority of our approach. However, our method is limited to text-based interaction. In the future, we aim to integrate additional modalities, such as images and depth maps, to enhance control and flexibility.

\textbf{Acknowledgments.} This work is partially supported by the National Natural Science Foundation of China (No.62406312), China National Postdoctoral Program for Innovative Talents (No.BX20240385) funded by China Postdoctoral Science Foundation.

\bibliography{ref}

\begin{thebibliography}{10}
\providecommand{\url}[1]{#1}
\csname url@samestyle\endcsname
\providecommand{\newblock}{\relax}
\providecommand{\bibinfo}[2]{#2}
\providecommand{\BIBentrySTDinterwordspacing}{\spaceskip=0pt\relax}
\providecommand{\BIBentryALTinterwordstretchfactor}{4}
\providecommand{\BIBentryALTinterwordspacing}{\spaceskip=\fontdimen2\font plus
\BIBentryALTinterwordstretchfactor\fontdimen3\font minus \fontdimen4\font\relax}
\providecommand{\BIBforeignlanguage}[2]{{%
\expandafter\ifx\csname l@#1\endcsname\relax
\typeout{** WARNING: IEEEtran.bst: No hyphenation pattern has been}%
\typeout{** loaded for the language `#1'. Using the pattern for}%
\typeout{** the default language instead.}%
\else
\language=\csname l@#1\endcsname
\fi
#2}}
\providecommand{\BIBdecl}{\relax}
\BIBdecl

\bibitem{ho2020denoising}
J.~Ho, A.~Jain, and P.~Abbeel, ``Denoising diffusion probabilistic models,'' \emph{Advances in neural information processing systems}, vol.~33, pp. 6840--6851, 2020.

\bibitem{podell2023sdxl}
D.~Podell, Z.~English, K.~Lacey, A.~Blattmann, T.~Dockhorn, J.~M{\"u}ller, J.~Penna, and R.~Rombach, ``Sdxl: Improving latent diffusion models for high-resolution image synthesis,'' \emph{arXiv preprint arXiv:2307.01952}, 2023.

\bibitem{rombach2022high}
R.~Rombach, A.~Blattmann, D.~Lorenz, P.~Esser, and B.~Ommer, ``High-resolution image synthesis with latent diffusion models,'' in \emph{Proceedings of the IEEE/CVF conference on computer vision and pattern recognition}, 2022, pp. 10\,684--10\,695.

\bibitem{clip}
A.~Radford, J.~W. Kim, C.~Hallacy, A.~Ramesh, G.~Goh, S.~Agarwal, G.~Sastry, A.~Askell, P.~Mishkin, J.~Clark \emph{et~al.}, ``Learning transferable visual models from natural language supervision,'' in \emph{International conference on machine learning}.\hskip 1em plus 0.5em minus 0.4em\relax PMLR, 2021, pp. 8748--8763.

\bibitem{yang2024clip}
C.~Yang, Z.~An, L.~Huang, J.~Bi, X.~Yu, H.~Yang, B.~Diao, and Y.~Xu, ``Clip-kd: An empirical study of clip model distillation,'' in \emph{Proceedings of the IEEE/CVF Conference on Computer Vision and Pattern Recognition}, 2024, pp. 15\,952--15\,962.

\bibitem{feng2024relational}
W.~Feng, C.~Yang, Z.~An, L.~Huang, B.~Diao, F.~Wang, and Y.~Xu, ``Relational diffusion distillation for efficient image generation,'' in \emph{Proceedings of the 32nd ACM International Conference on Multimedia}, 2024, pp. 205--213.

\bibitem{betker2023improving}
J.~Betker, G.~Goh, L.~Jing, T.~Brooks, J.~Wang, L.~Li, L.~Ouyang, J.~Zhuang, J.~Lee, Y.~Guo \emph{et~al.}, ``Improving image generation with better captions,'' \emph{Computer Science. https://cdn. openai. com/papers/dall-e-3. pdf}, vol.~2, no.~3, p.~8, 2023.

\bibitem{saharia2022photorealistic}
C.~Saharia, W.~Chan, S.~Saxena, L.~Li, J.~Whang, E.~L. Denton, K.~Ghasemipour, R.~Gontijo~Lopes, B.~Karagol~Ayan, T.~Salimans \emph{et~al.}, ``Photorealistic text-to-image diffusion models with deep language understanding,'' \emph{Advances in neural information processing systems}, vol.~35, pp. 36\,479--36\,494, 2022.

\bibitem{li2024hunyuan}
Z.~Li, J.~Zhang, Q.~Lin, J.~Xiong, Y.~Long, X.~Deng, Y.~Zhang, X.~Liu, M.~Huang, Z.~Xiao \emph{et~al.}, ``Hunyuan-dit: A powerful multi-resolution diffusion transformer with fine-grained chinese understanding,'' \emph{arXiv preprint arXiv:2405.08748}, 2024.

\bibitem{chen2023pixart}
J.~Chen, J.~Yu, C.~Ge, L.~Yao, E.~Xie, Y.~Wu, Z.~Wang, J.~Kwok, P.~Luo, H.~Lu \emph{et~al.}, ``Pixart-$\alpha$: Fast training of diffusion transformer for photorealistic text-to-image synthesis,'' \emph{arXiv preprint arXiv:2310.00426}, 2023.

\bibitem{wang2024gra}
J.~Wang, Y.~Pu, Y.~Han, J.~Guo, Y.~Wang, X.~Li, and G.~Huang, ``Gra: Detecting oriented objects through group-wise rotating and attention,'' \emph{arXiv preprint arXiv:2403.11127}, 2024.

\bibitem{glide}
T.~A. Halgren, R.~B. Murphy, R.~A. Friesner, H.~S. Beard, L.~L. Frye, W.~T. Pollard, and J.~L. Banks, ``Glide: a new approach for rapid, accurate docking and scoring. 2. enrichment factors in database screening,'' \emph{Journal of medicinal chemistry}, vol.~47, no.~7, pp. 1750--1759, 2004.

\bibitem{transformer}
A.~Vaswani, ``Attention is all you need,'' \emph{Advances in Neural Information Processing Systems}, 2017.

\bibitem{clipscore}
J.~Hessel, A.~Holtzman, M.~Forbes, R.~L. Bras, and Y.~Choi, ``Clipscore: A reference-free evaluation metric for image captioning,'' \emph{arXiv preprint arXiv:2104.08718}, 2021.

\bibitem{zhang2023adding}
L.~Zhang, A.~Rao, and M.~Agrawala, ``Adding conditional control to text-to-image diffusion models,'' in \emph{Proceedings of the IEEE/CVF International Conference on Computer Vision}, 2023, pp. 3836--3847.

\bibitem{li2023gligen}
Y.~Li, H.~Liu, Q.~Wu, F.~Mu, J.~Yang, J.~Gao, C.~Li, and Y.~J. Lee, ``Gligen: Open-set grounded text-to-image generation,'' in \emph{Proceedings of the IEEE/CVF Conference on Computer Vision and Pattern Recognition}, 2023, pp. 22\,511--22\,521.

\bibitem{bar2023multidiffusion}
O.~Bar-Tal, L.~Yariv, Y.~Lipman, and T.~Dekel, ``Multidiffusion: Fusing diffusion paths for controlled image generation,'' 2023.

\bibitem{wang2024taming}
J.~Wang, J.~Pu, Z.~Qi, J.~Guo, Y.~Ma, N.~Huang, Y.~Chen, X.~Li, and Y.~Shan, ``Taming rectified flow for inversion and editing,'' \emph{arXiv preprint arXiv:2411.04746}, 2024.

\bibitem{zhu2024multibooth}
C.~Zhu, K.~Li, Y.~Ma, C.~He, and L.~Xiu, ``Multibooth: Towards generating all your concepts in an image from text,'' \emph{arXiv preprint arXiv:2404.14239}, 2024.

\bibitem{dino}
H.~Zhang, F.~Li, S.~Liu, L.~Zhang, H.~Su, J.~Zhu, L.~M. Ni, and H.-Y. Shum, ``Dino: Detr with improved denoising anchor boxes for end-to-end object detection,'' \emph{arXiv preprint arXiv:2203.03605}, 2022.

\bibitem{yang2022cross}
C.~Yang, H.~Zhou, Z.~An, X.~Jiang, Y.~Xu, and Q.~Zhang, ``Cross-image relational knowledge distillation for semantic segmentation,'' in \emph{Proceedings of the IEEE/CVF Conference on Computer Vision and Pattern Recognition}, 2022, pp. 12\,319--12\,328.

\bibitem{li2024sglp}
Y.~Li, Y.~Lu, Z.~Dong, C.~Yang, Y.~Chen, and J.~Gou, ``Sglp: A similarity guided fast layer partition pruning for compressing large deep models,'' \emph{arXiv preprint arXiv:2410.14720}, 2024.

\bibitem{li2024comae}
Y.~Li, Q.~Long, Y.~Zhou, N.~Cao, S.~Liu, F.~Zheng, Z.~Zhu, Z.~Ning, M.~Xiao, X.~Wang \emph{et~al.}, ``Comae: Comprehensive attribute exploration for zero-shot hashing,'' \emph{arXiv preprint arXiv:2402.16424}, 2024.

\bibitem{ma2024followyourpose}
Y.~Ma, Y.~He, X.~Cun, X.~Wang, S.~Chen, X.~Li, and Q.~Chen, ``Follow your pose: Pose-guided text-to-video generation using pose-free videos,'' in \emph{Proceedings of the AAAI Conference on Artificial Intelligence}, vol.~38, no.~5, 2024, pp. 4117--4125.

\bibitem{ma2024followyourclick}
Y.~Ma, Y.~He, H.~Wang, A.~Wang, C.~Qi, C.~Cai, X.~Li, Z.~Li, H.-Y. Shum, W.~Liu \emph{et~al.}, ``Follow-your-click: Open-domain regional image animation via short prompts,'' \emph{arXiv preprint arXiv:2403.08268}, 2024.

\bibitem{ma2022visual}
Y.~Ma, Y.~Wang, Y.~Wu, Z.~Lyu, S.~Chen, X.~Li, and Y.~Qiao, ``Visual knowledge graph for human action reasoning in videos,'' in \emph{Proceedings of the 30th ACM International Conference on Multimedia}, 2022, pp. 4132--4141.

\bibitem{ma2023magicstick}
Y.~Ma, X.~Cun, Y.~He, C.~Qi, X.~Wang, Y.~Shan, X.~Li, and Q.~Chen, ``Magicstick: Controllable video editing via control handle transformations,'' \emph{arXiv preprint arXiv:2312.03047}, 2023.

\bibitem{ma2024followyouremoji}
Y.~Ma, H.~Liu, H.~Wang, H.~Pan, Y.~He, J.~Yuan, A.~Zeng, C.~Cai, H.-Y. Shum, W.~Liu \emph{et~al.}, ``Follow-your-emoji: Fine-controllable and expressive freestyle portrait animation,'' \emph{arXiv preprint arXiv:2406.01900}, 2024.

\bibitem{chen2024follow}
Q.~Chen, Y.~Ma, H.~Wang, J.~Yuan, W.~Zhao, Q.~Tian, H.~Wang, S.~Min, Q.~Chen, and W.~Liu, ``Follow-your-canvas: Higher-resolution video outpainting with extensive content generation,'' \emph{arXiv preprint arXiv:2409.01055}, 2024.

\bibitem{wang2024cove}
J.~Wang, Y.~Ma, J.~Guo, Y.~Xiao, G.~Huang, and X.~Li, ``Cove: Unleashing the diffusion feature correspondence for consistent video editing,'' \emph{arXiv preprint arXiv:2406.08850}, 2024.

\bibitem{zhu2024instantswap}
C.~Zhu, K.~Li, Y.~Ma, L.~Tang, C.~Fang, C.~Chen, Q.~Chen, and X.~Li, ``Instantswap: Fast customized concept swapping across sharp shape differences,'' \emph{arXiv preprint arXiv:2412.01197}, 2024.

\bibitem{feng2024dit4edit}
K.~Feng, Y.~Ma, B.~Wang, C.~Qi, H.~Chen, Q.~Chen, and Z.~Wang, ``Dit4edit: Diffusion transformer for image editing,'' \emph{arXiv preprint arXiv:2411.03286}, 2024.

\bibitem{t5}
C.~Raffel, N.~Shazeer, A.~Roberts, K.~Lee, S.~Narang, M.~Matena, Y.~Zhou, W.~Li, and P.~J. Liu, ``Exploring the limits of transfer learning with a unified text-to-text transformer,'' \emph{Journal of machine learning research}, vol.~21, no. 140, pp. 1--67, 2020.

\bibitem{db}
N.~Ruiz, Y.~Li, V.~Jampani, Y.~Pritch, M.~Rubinstein, and K.~Aberman, ``Dreambooth: Fine tuning text-to-image diffusion models for subject-driven generation,'' in \emph{Proceedings of the IEEE/CVF conference on computer vision and pattern recognition}, 2023, pp. 22\,500--22\,510.

\bibitem{ipada}
H.~Ye, J.~Zhang, S.~Liu, X.~Han, and W.~Yang, ``Ip-adapter: Text compatible image prompt adapter for text-to-image diffusion models,'' \emph{arXiv preprint arXiv:2308.06721}, 2023.

\bibitem{ma2024discrepancy}
Z.~Ma, Y.~Li, Y.~Luo, X.~Luo, J.~Li, C.~Chen, X.-S. Hua, and G.~Lu, ``Discrepancy and structure-based contrast for test-time adaptive retrieval,'' \emph{IEEE Transactions on Multimedia}, 2024.

\bibitem{lu2024generic}
Y.~Lu, Y.~Zhu, Y.~Li, D.~Xu, Y.~Lin, Q.~Xuan, and X.~Yang, ``A generic layer pruning method for signal modulation recognition deep learning models,'' \emph{arXiv preprint arXiv:2406.07929}, 2024.

\bibitem{peebles2023scalable}
W.~Peebles and S.~Xie, ``Scalable diffusion models with transformers,'' in \emph{Proceedings of the IEEE/CVF International Conference on Computer Vision}, 2023, pp. 4195--4205.

\end{thebibliography}
\bibliographystyle{IEEEtran}

\end{document}